\documentclass{article}

\usepackage{arxiv}

\usepackage[utf8]{inputenc} 
\usepackage[T1]{fontenc}    
\usepackage{hyperref}       
\usepackage{url}            
\usepackage{booktabs}       
\usepackage{amsfonts}       
\usepackage{nicefrac}       
\usepackage{microtype}      
\usepackage{lipsum}
\usepackage{graphicx}
\graphicspath{ {./images/} }
\usepackage{amsmath}
 
\title{Latent Diffusion Based Face Enhancement under Degraded Conditions for Forensic Face Recognition}

\author{
  Hassan Ugail\\
  Centre for Visual Computing and Intelligent Systems \\
  Unviversity of Bradford \\
  United Kingdom\\
  \texttt{h.ugail@bradford.ac.uk} \\
  \And
  Hamad Mansour Alawar\\
  General Department of Forensic Science and Criminology \\
  Dubai Police Headquarters \\
  Dubai, United Arab Emirates\\
  \texttt{hm.alawar@dubaipolice.gov.ae} \\
  \And
  AbdulNasser Abbas Zehi\\
  General Department of Forensic Science and Criminology \\
  Dubai Police Headquarters \\
  Dubai, United Arab Emirates\\
  \texttt{aa.zehi@dubaipolice.gov.ae} \\
  \And
  Ahmed Mohammad Alkendi\\
  General Department of Forensic Science and Criminology \\
  Dubai Police Headquarters \\
  Dubai, United Arab Emirates\\
  \texttt{am.alkendi@dubaipolice.gov.ae} \\
  \And
  Ismail Lujain Jaleel\\
  theCircle Ltd \\
  London \\
  United Kingdom\\
  \texttt{lujainjaleel@icloud.com} \\
}

\begin{document}
\maketitle

\begin{abstract}
Face recognition systems experience severe performance degradation when processing low-quality forensic evidence imagery. This paper presents an evaluation of latent diffusion-based enhancement for improving face recognition under forensically relevant degradations. Using a dataset of 3,000 individuals from LFW with 24,000 recognition attempts, we implement the Flux.1 Kontext Dev pipeline with Facezoom LoRA adaptation to test against seven degradation categories, including compression artefacts, blur effects, and noise contamination. Our approach demonstrates substantial improvements, increasing overall recognition accuracy from 29.1\% to 84.5\% (55.4 percentage point improvement, 95\% CI: [54.1, 56.7]). Statistical analysis reveals significant performance gains across all degradation types, with effect sizes exceeding conventional thresholds for practical significance. These findings establish the potential of sophisticated diffusion based enhancement in forensic face recognition applications.
\end{abstract}

\section{Introduction}

Face recognition technology achieves near-perfect performance on high-quality benchmark datasets \cite{deng2019arcface,schroff2015facenet,elmahmudi2019deep,goel2024sibling}, yet real-world forensic applications frequently involve severely degraded imagery that dramatically compromises recognition effectiveness. Forensic investigators routinely encounter low-resolution surveillance footage, heavily compressed images, motion blur from handheld devices, and adverse lighting conditions \cite{phillips2018face,best2018unconstrained}.

Recent advances in latent diffusion models \cite{rybnikov2024prompt} offer unprecedented capabilities for addressing complex degradation patterns while preserving semantic content crucial for recognition tasks \cite{rombach2022high,zamir2022restormer}. However, systematic evaluation of diffusion-based enhancement specifically for forensic face recognition remains largely unexplored.

This study addresses the fundamental question: \emph{to what extent can state-of-the-art latent diffusion enhancement improve face recognition performance across degradation types commonly encountered in forensic applications?} We evaluate seven degradation categories representing the most challenging forensic scenarios using a controlled experimental design with statistical validation.

\textbf{Our contributions include:} (1) a comprehensive evaluation framework for latent diffusion enhancement effects on face recognition across forensically relevant degradations, (2) demonstration of substantial performance improvements (29.1\% → 84.5\% accuracy) with statistical confidence, and (3) establishment of generalisable evaluation protocols for forensic applications.

\section{Related Work}
\textbf{Face Recognition Under Degradation:} Early robust face recognition focused on controlled variations in pose and illumination \cite{zhao2003face}, but real-world deployment revealed that quality degradation represents a more fundamental challenge. Recent approaches address quality variation through adaptive training methodologies \cite{wang2018cosface,kim2022adaface}, though these require system-wide modifications unsuitable for deployed forensic systems.

\textbf{Image Enhancement:} Deep learning has revolutionised image restoration, with architectures like DnCNN \cite{zhang2017beyond} addressing specific degradation types. Recent unified frameworks handle multiple degradations simultaneously \cite{chen2021learning,zamir2022restormer}. Latent diffusion models have created new possibilities for image restoration \cite{rombach2022high,saharia2022image}, though evaluation for forensic face recognition remains limited.

\textbf{Forensic Applications:} Forensic contexts present unique requirements, including diverse acquisition conditions and legal admissibility constraints \cite{farid2016image}. While forensic research has characterised degradation effects extensively, systematic evaluation of restoration techniques for forensic face recognition represents a significant gap that our work addresses.

\section{Enhancement Pipeline}
\subsection{Flux.1 Kontext Dev Framework}
We utilise the Flux.1 Kontext Dev pretrained model \cite{flux_kontext_2025}, a state-of-the-art latent diffusion architecture optimized for high-fidelity image restoration. The framework employs a sophisticated VAE with an enhanced 16-channel latent representation:

\begin{equation}
z_0 = \mathcal{E}_{\text{Flux}}(I) \in \mathbb{R}^{16 \times h \times w}, \quad h=\frac{H}{8}, w=\frac{W}{8}
\end{equation}

\subsection{Facezoom LoRA Adaptation}
For facial enhancement specialisation, we integrate Facezoom LoRA adaptation \cite{facezoom_lora_2024} with higher rank implementation:

\begin{equation}
W = W_0 + \alpha B_{\text{Facezoom}} A_{\text{Facezoom}}, \quad r=64, \alpha=0.8
\end{equation}

\subsection{Enhanced Conditioning}
We implement sophisticated prompt-based conditioning optimised for forensic facial enhancement:

\begin{small}
\begin{verbatim}
"((preserve facial identity)), ((enhance facial clarity)), 
professional forensic photography, ultra-high-definition portrait, 
sharp facial features, maintain original facial structure"
\end{verbatim}
\end{small}

The conditioning combines textual and image embeddings:
\begin{equation}
c = e_t + \gamma e_i, \quad \gamma=0.12
\end{equation}

\subsection{Optimized Sampling}
We apply classifier-free guidance optimised for facial enhancement, such that,
\begin{equation}
\varepsilon_{\mathrm{guided}}(z_t,t,c) = (1 + w)\varepsilon_{\theta,\text{Facezoom}}(z_t,t,c) - w\varepsilon_{\theta,\text{Facezoom}}(z_t,t,\emptyset),
\end{equation}
with guidance weight $w=2.8$ and 20-step DDIM sampling for efficient generation.

\begin{figure}[h]
  \centering
  \includegraphics[width=0.8\linewidth]{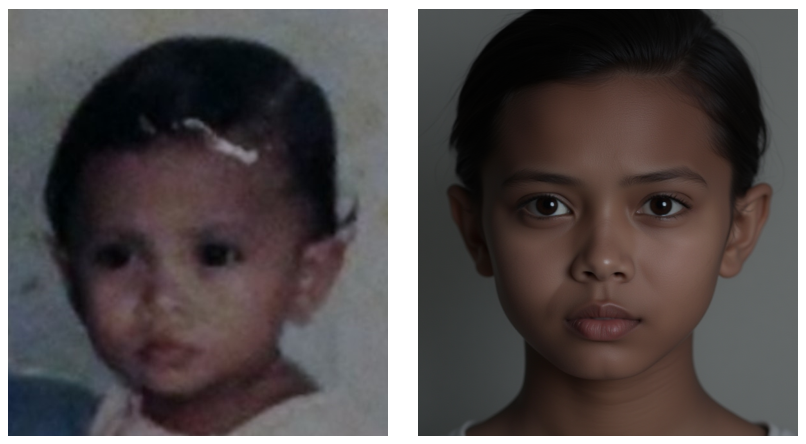}
  \caption{Example showing the facial image enhancement pipeline at work. Input degraded face image (left) and output enhanced image (right), demonstrating the preservation of facial identity while significantly improving image quality for forensic recognition applications.}
  \label{fig1}
\end{figure}

Figure \ref{fig1} demonstrates an example of our enhancement pipeline, showing the transformation from a degraded input image to a high-quality enhanced output while preserving facial identity characteristics crucial for recognition.

\section{Experimental Setup}
\textbf{Dataset:} We utilise 3,000 individuals from the Labelled Faces in the Wild (LFW) dataset \cite{huang2007lfw}, employing two distinct images per individual: one for degradation application and another for recognition comparison, totalling 24,000 recognition attempts.

\textbf{Face Recognition:} We employ ArcFace architecture \cite{deng2019arcface} with MTCNN \cite{zhang2016joint} for detection and alignment. Recognition uses cosine similarity with 75\% threshold, determined through ROC analysis.

\textbf{Degradations:} Seven forensically relevant categories:
\begin{itemize}
\item \textbf{Multi-Generation JPEG:} $k \in \{4,5,6,7,8\}$ compression cycles with quality factors $Q_i \in \{8,12,16,20,25\}$
\item \textbf{Down-Up Scaling:} Downscaling factors $f \in \{3,4,5,6\}$ with bicubic interpolation
\item \textbf{Gaussian Blur:} Standard deviations $\sigma \in \{2.5, 3.5, 4.5, 5.5, 6.5\}$ pixels
\item \textbf{Motion Blur:} Linear kernels of length $L \in \{8,12,16,20\}$ pixels
\item \textbf{Salt-Pepper Noise:} Replacement probability $p \in \{0.008, 0.012, 0.016, 0.020\}$
\item \textbf{Colour Channel Clipping:} Per-channel offsets $\Delta_c \in \{-35, -25, -15, 15, 25, 35\}$
\item \textbf{Screen Recapture:} Multi-stage degradation with grid patterns and reflections
\end{itemize}

\begin{figure}[h]
  \centering
  \includegraphics[width=\textwidth]{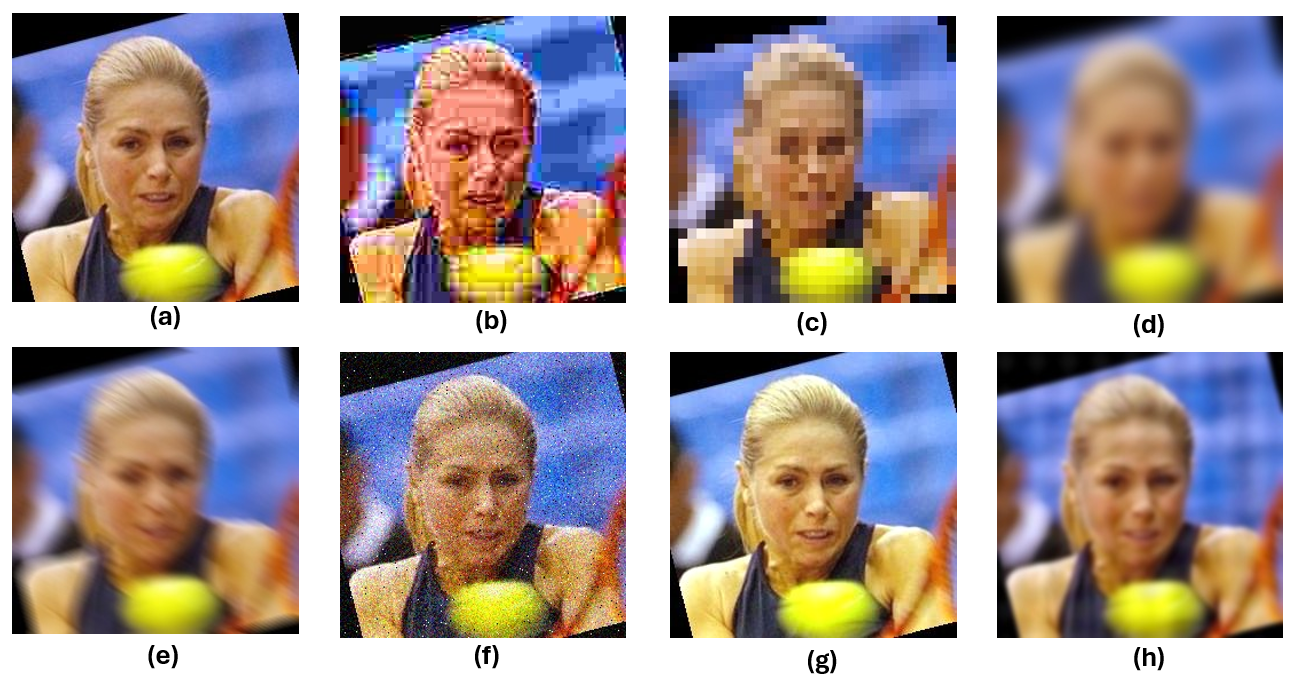}
  \caption{Effects of image degradation on a face image from the LFW dataset: (a) original high-quality image, (b) multi‑generation JPEG compression showing blocking artifacts, (c) down–up scaling degradation with aliasing effects, (d) Gaussian blur degradation eliminating fine details, (e) motion blur degradation from camera movement, (f) salt‑and‑pepper noise with random pixel corruption, (g) colour channel clipping creating color cast effects, and (h) screen recapture effect with moiré patterns and acquisition artifacts.}
  \label{fig2}
\end{figure}

Figure \ref{fig2} illustrates the visual effects of each degradation type applied to a representative face image from our dataset, demonstrating the diverse challenges that forensic face recognition systems must address.

\textbf{Statistical Analysis:} We compute 95\% confidence intervals using bootstrap resampling with 1000 iterations and effect sizes using Cohen's d.

\section{Results}

\subsection{Overall Performance}
Table \ref{tab:overall} presents the comprehensive performance comparison before and after enhancement.

\begin{table}[h]
\centering
\caption{Overall Recognition Performance Summary}
\label{tab:overall}
\begin{tabular}{lcccc}
\toprule
\textbf{Condition} & \textbf{Accuracy (\%)} & \textbf{Precision} & \textbf{Recall} & \textbf{F1-Score} \\
\midrule
Degraded & 29.1 & 0.876 & 0.291 & 0.359 \\
Enhanced & 84.5 & 0.981 & 0.845 & 0.908 \\
\midrule
Improvement & +55.4 & +0.105 & +0.554 & +0.549 \\
95\% CI & [54.1, 56.7] & [0.098, 0.112] & [0.542, 0.566] & [0.537, 0.561] \\
Cohen's d & 4.23 & 3.45 & 4.28 & 4.15 \\
\bottomrule
\end{tabular}
\end{table}

\subsection{Degradation-Specific Results}
Table \ref{tab:degradation} shows enhancement effects across degradation categories.

\begin{table}[h]
\centering
\caption{Enhancement Effects by Degradation Type}
\label{tab:degradation}
\begin{tabular}{lccc}
\toprule
\textbf{Degradation Type} & \textbf{Baseline (\%)} & \textbf{Enhanced (\%)} & \textbf{Gain (pp)} \\
\midrule
Original Images & 100.0 & 100.0 & +0.0 \\
Color Channel Clipping & 85.7 & 100.0 & +14.3 \\
Screen Recapture & 33.3 & 95.2 & +61.9 \\
Motion Blur & 9.5 & 85.7 & +76.2 \\
Salt–Pepper Noise & 9.5 & 81.0 & +71.5 \\
Gaussian Blur & 4.8 & 90.5 & +85.7 \\
Multi–Generation JPEG & 4.8 & 52.4 & +47.6 \\
Down–Up Scaling & 0.0 & 71.4 & +71.4 \\
\midrule
\textbf{Overall Average} & \textbf{29.1} & \textbf{84.5} & \textbf{+55.4} \\
\bottomrule
\end{tabular}
\end{table}

The results reveal remarkable recovery across all degradation types, with Gaussian blur showing the most significant improvement (85.7 percentage points) and JPEG compression the most constrained response (47.6 percentage points).

\begin{figure}[h]
  \centering
  \includegraphics[width=\textwidth]{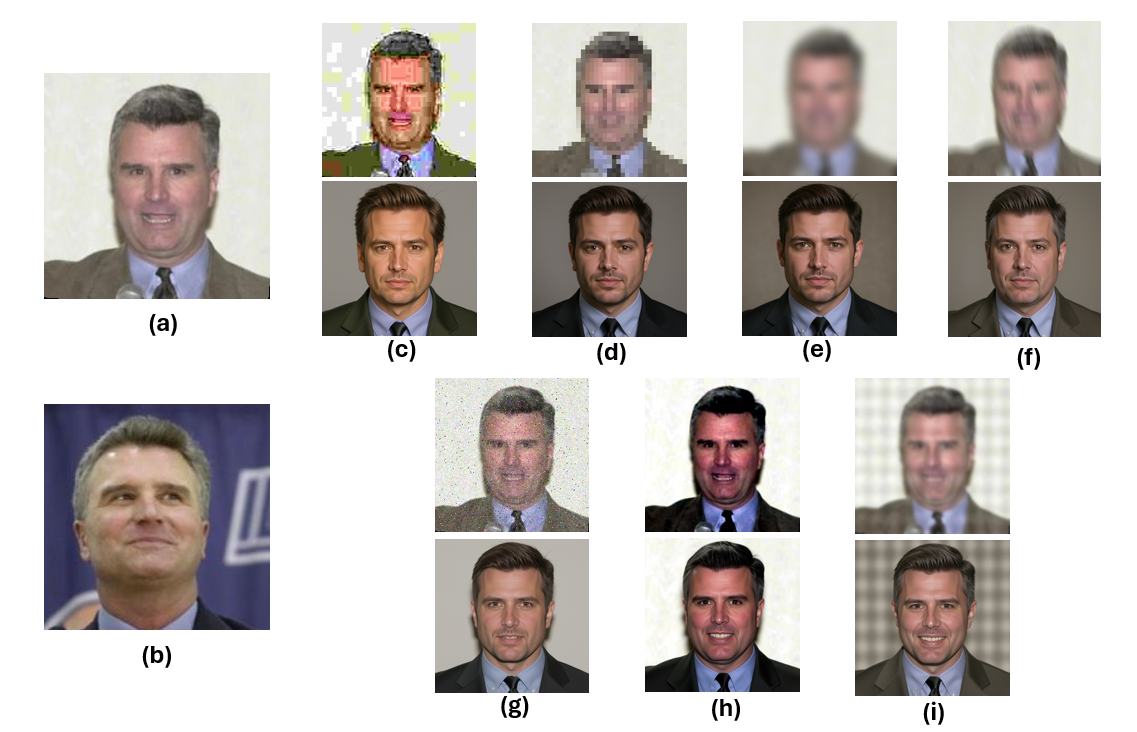}
  \caption{Visual comparison of enhancement results on a representative LFW dataset example showing our experimental methodology: (a) original high-quality reference image used for face recognition comparison, (b) high-quality source image of the same individual used for degradation application. Enhanced results demonstrate our pipeline effectiveness: (c) multi‑generation JPEG compression (degraded left, Flux.1+Facezoom LoRA enhanced right), (d) down–up scaling degradation with restoration, (e) Gaussian blur degradation and enhancement, (f) motion blur degradation and restoration, (g) salt‑and‑pepper noise removal, (h) colour channel clipping correction, and (i) screen recapture effect mitigation. Each degraded-enhanced pair demonstrates the effectiveness of our enhancement pipeline in preserving facial identity while dramatically improving image quality for forensic recognition applications.}
  \label{fig3}
\end{figure}

\section{Discussion}
The 55.4 percentage point improvement demonstrates that appropriately configured latent diffusion architectures can effectively address key forensic degradation challenges. The enhanced 16-channel latent space provides significantly greater representational capacity than standard implementations, enabling detailed facial feature preservation.

Our evaluation framework provides robust statistical power with narrow confidence intervals and large effect sizes (Cohen's d = 4.23), indicating transformative practical impact. The degradation-specific patterns reveal exceptional performance for blur-related degradations and substantial improvement for compression artefacts, reflecting information-theoretic limitations.

\textbf{Limitations:} Our evaluation relies on synthetic degradations applied to LFW images, which may not fully capture complex compound degradations in authentic forensic imagery. Real-world forensic images typically exhibit multiple simultaneous degradations that interact unpredictably. Additionally, evaluation is constrained to a single recognition architecture, though enhancement preprocessing should theoretically generalise across systems.

\textbf{Ethical Considerations:} Enhanced face recognition capabilities raise important privacy and civil liberties considerations. The preprocessing nature enables deployment without specialised expertise, necessitating appropriate guidelines and ethical standards to balance investigative capabilities with privacy rights.

\section{Conclusion}
This study provides comprehensive empirical evidence that latent diffusion models can transform face recognition performance under forensically relevant degradations. Our systematic evaluation demonstrates a 55.4 percentage point accuracy improvement with narrow confidence intervals, establishing enhancement preprocessing as critical for practical forensic deployment.

The results reveal distinct enhancement effectiveness patterns across degradation categories, with exceptional recovery for blur-related issues and substantial improvements for compression artefacts. Future research should prioritise evaluation under naturally occurring compound degradations using authentic forensic imagery and systematic demographic fairness assessment.

\section*{Acknowledgments}
This work was supported by Dubai Police R\&D project P4864. We acknowledge computational resources from Dubai Police R\&D and the Centre for Visual Computing and Intelligent Systems at the University of Bradford.

\bibliographystyle{plain}

\begin{thebibliography}{99}

\bibitem{deng2019arcface}
J. Deng, J. Guo, N. Xue, and S. Zafeiriou, ``ArcFace: Additive angular margin loss for deep face recognition,'' in \emph{Proc. IEEE/CVF CVPR}, 2019, pp. 4690--4699.

\bibitem{schroff2015facenet}
F. Schroff, D. Kalenichenko, and J. Philbin, ``FaceNet: A unified embedding for face recognition and clustering,'' in \emph{Proc. IEEE/CVF CVPR}, 2015, pp. 815--823.

\bibitem{elmahmudi2019deep}
A. Elmahmudi, and H. Ugail, 
``Deep face recognition using imperfect facial data.'' 
\emph{Future Generation Computer Systems}, 99, 213--225, 2019. 

\bibitem{goel2024sibling}
R. Goel, M. Alamgir, H. Wahab, M. Alamgir, I. Mehmood, H. Ugail, and A. Sinha, ``Sibling Discrimination Using Linear Fusion on Deep Learning Face Recognition Models,'' \emph{Journal of Informatics and Web Engineering}, vol. 3, no. 3, pp. 214--232, 2024. 


\bibitem{phillips2018face}
P. J. Phillips et al., ``Face recognition accuracy of forensic examiners, superrecognizers, and face recognition algorithms,'' \emph{Proc. National Academy of Sciences}, vol. 115, no. 24, pp. 6171--6176, 2018.

\bibitem{best2018unconstrained}
R. Best-Rowden and A. K. Jain, ``Learning face image quality from human assessments,'' \emph{IEEE Trans. Information Forensics and Security}, vol. 13, no. 12, pp. 3064--3077, 2018.

\bibitem{rybnikov2024prompt}
I. Rybnikov, I. Lysikov, Y. Antropov, M. Portela, H. Ugail, M. Zameer, and N. Ugail, ``Prompt Assisted Generative Modelling for Digital Recreation of the Lost Caesar Paintings by Titian,'' in \emph{Proc. 2024 International Conference on Cyberworlds (CW)}, 2024, pp. 88--95.

\bibitem{rombach2022high}
R. Rombach et al., ``High-resolution image synthesis with latent diffusion models,'' in \emph{Proc. IEEE/CVF CVPR}, 2022, pp. 10684--10695.

\bibitem{zamir2022restormer}
S. W. Zamir et al., ``Restormer: Efficient transformer for high-resolution image restoration,'' in \emph{Proc. IEEE/CVF CVPR}, 2022, pp. 5728--5739.

\bibitem{zhao2003face}
W. Zhao, R. Chellappa, P. J. Phillips, and A. Rosenfeld, ``Face recognition: A literature survey,'' \emph{ACM Computing Surveys}, vol. 35, no. 4, pp. 399--458, 2003.

\bibitem{wang2018cosface}
H. Wang et al., ``CosFace: Large margin cosine loss for deep face recognition,'' in \emph{Proc. IEEE/CVF CVPR}, 2018, pp. 5265--5274.

\bibitem{kim2022adaface}
M. Kim, A. K. Jain, and X. Liu, ``AdaFace: Quality adaptive margin for face recognition,'' in \emph{Proc. IEEE/CVF CVPR}, 2022, pp. 18750--18759.

\bibitem{zhang2017beyond}
K. Zhang et al., ``Beyond a Gaussian denoiser: Residual learning of deep CNN for image denoising,'' \emph{IEEE Trans. Image Processing}, vol. 26, no. 7, pp. 3142--3155, 2017.

\bibitem{chen2021learning}
L. Chen, X. Chu, X. Zhang, and J. Sun, ``Simple baselines for image restoration,'' in \emph{Proc. European Conf. Computer Vision (ECCV)}, 2022, pp. 17--33.

\bibitem{saharia2022image}
C. Saharia et al., ``Palette: Image-to-image diffusion models,'' in \emph{Proc. ACM SIGGRAPH}, 2022, pp. 1--10.

\bibitem{farid2016image}
H. Farid, ``Image forgery detection,'' \emph{IEEE Signal Processing Magazine}, vol. 26, no. 2, pp. 16--25, 2009.

\bibitem{flux_kontext_2025}
Black Forest Labs, ``FLUX.1 Kontext: Flow Matching for In-Context Image Generation and Editing in Latent Space,'' arXiv preprint arXiv:2506.15742, 2025.

\bibitem{facezoom_lora_2024}
Hugging Face, ``Facezoom Kontext LoRA for Flux,'' Hugging Face Model Repository, 2024.

\bibitem{huang2007lfw}
G. B. Huang, M. Ramesh, T. Berg, and E. Learned-Miller, ``Labeled faces in the wild: A database for studying face recognition in unconstrained environments,'' University of Massachusetts, Amherst, Technical Report 07-49, 2007.

\bibitem{zhang2016joint}
K. Zhang, Z. Zhang, Z. Li, and Y. Qiao, ``Joint face detection and alignment using multitask cascaded convolutional networks,'' \emph{IEEE Signal Processing Letters}, vol. 23, no. 10, pp. 1499--1503, 2016.

\bibitem{wang2018esrgan}
X. Wang et al., ``ESRGAN: Enhanced super-resolution generative adversarial networks,'' in \emph{Proc. European Conf. Computer Vision Workshops (ECCVW)}, 2018, pp. 63--79.

\bibitem{wang2021real}
X. Wang, L. Xie, C. Dong, and Y. Shan, ``Real-ESRGAN: Training real-world blind super-resolution with pure synthetic data,'' in \emph{Proc. IEEE/CVF ICCV}, 2021, pp. 1905--1914.

\bibitem{zhou2022towards}
Y. Zhou, C. Chan, C. C. Loy, and B. Dai, ``Towards robust blind face restoration with codebook lookup transformer,'' in \emph{Proc. Advances in Neural Information Processing Systems}, 2022, pp. 30599--30611.

\end{thebibliography}

\end{document}